\PassOptionsToPackage{table,xcdraw}{xcolor}
\documentclass[sigconf,nonacm]{acmart}

\copyrightyear{2024}
\acmYear{2024}
\setcopyright{rightsretained}
\acmConference[CHI EA '24]{Extended Abstracts of the CHI Conference on Human Factors in Computing Systems}{May 11--16, 2024}{Honolulu, HI, USA}
\acmBooktitle{Extended Abstracts of the CHI Conference on Human Factors in Computing Systems (CHI EA '24), May 11--16, 2024, Honolulu, HI, USA}
\acmDOI{10.1145/3613905.3651007}
\acmISBN{979-8-4007-0331-7/24/05}

\usepackage{multirow}
\usepackage[utf8]{inputenc}
\usepackage{wrapfig}
\usepackage{graphicx}
\usepackage[table,xcdraw]{xcolor}
\usepackage[toc,page]{appendix}

\begin{document}

%%
%% The "title" command has an optional parameter,
%% allowing the author to define a "short title" to be used in page headers.
\title{Understanding the Dataset Practitioners Behind Large Language Model Development}

\settopmatter{authorsperrow=3, printfolios=true}

\author{Crystal Qian}
\authornote{Both authors contributed equally to this research.}
\orcid{0000-0001-7716-7245}
\email{cjqian@google.com}
\affiliation{%
  \institution{Google Research}
  \city{New York City}
  \state{NY}
  \country{USA}
}

\author{Emily Reif}
\email{ereif@google.com}
\authornotemark[1]
\orcid{0000-0003-3572-6234}
\affiliation{%
  \institution{Google Research}
  \city{Seattle}
  \state{WA}
  \country{USA}
}

\author{Minsuk Kahng}
\orcid{0000-0002-0291-6026}
\email{kahng@google.com}
\affiliation{%
  \institution{Google Research}
  \city{Atlanta}
  \state{GA}
  \country{USA}
}

%%
%% The abstract is a short summary of the work to be presented in the
%% article.
\begin{abstract}
As large language models (LLMs) become more advanced and impactful, it is increasingly important to scrutinize the data that they rely upon and produce. What is it to be a dataset practitioner doing this work? We approach this in two parts: first, we define the role of ``dataset practitioners'' by performing a retrospective analysis on the responsibilities of teams contributing to LLM development at a technology company, Google. Then, we conduct semi-structured interviews with a cross-section of these practitioners (N=10). We find that although data quality is a top priority, there is little consensus around what data quality is and how to evaluate it. Consequently, practitioners either rely on their own intuition or write custom code to evaluate their data. We discuss potential reasons for this phenomenon and opportunities for alignment.
\end{abstract}




%%
%% This command processes the author and affiliation and title
%% information and builds the first part of the formatted document.
\maketitle

\section{Introduction}
As the state-of-the-art for large language models (LLMs) advances \cite{openai2023gpt,team2023gemini}, the field of relevant data analysis is rapidly evolving. Because the data used and produced by LLMs is largely unstructured, traditional statistical analyses are insufficient for rigorous evaluation  \cite{wach2023,sambasivan, brown}. Furthermore, as applications of these LLMs become more widely adopted and impactful \cite{abdullah2022, baldassarre}, there is a deeper need to qualitatively understand these datasets; for instance, to mitigate sociological biases, ensure safe outputs, and minimize harm.

We aim to identify the needs and challenges of those who want to understand unstructured, text-based datasets for LLM development: a group that we define as \textbf{dataset practitioners}. To develop this definition, we perform a retrospective analysis within Google, a technology company that is developing LLMs. We then conduct semi-structured interviews with a cross-section of practitioners (N=10) to better understand their workflows, tools, and challenges.

We find that practitioners increasingly prioritize data quality; however, there is no consensus on what constitutes ``high quality'' data. Despite the HCI and visualization researchers' active efforts to deliver relevant sensemaking methods and tools, data practitioners in aggregate do not appear to be adopting these solutions, instead relying either on cursory visual inspection of spreadsheets or custom analyses logic in notebooks to understand their data. There is demand for frameworks, consensus, and tooling in this space that is not being met. We discuss hypotheses for this observed phenomenon, and conclude with opportunities for further research and alignment.

\section{Related Work}\label{sec:related_work}

\subsection{Analyzing Analyzers}\label{sec:related_analyzers} As data science has grown as a discipline, so have the amount of analyses \cite{harris2013analyzing, crisan}, surveys \cite{zhang2020}, and interviews \cite{wang2019} performed to capture the role of those who do this work. 

Some notable highlights include Kandel et al. \cite{kandel}, which classifies the emerging role of the \textit{data analysts} across different industrial sectors, such as healthcare and retail. Muller et al. \cite{muller} interviewed \textit{data scientists} at IBM to capture different approaches to their work, and  Crisan et al. \cite{crisan} creates a taxonomy of job roles across \textit{data workers}, such as \textit{moonlighters, generalists} or \textit{evangelists}.

Across these studies, the definitions of \textit{data analysts} or \textit{data workers} satisfy the breadth of work that we aim to capture in this inquiry. \textit{Data scientist} is too narrow for our population. It does not encompass the specific challenges introduced by the new LLM-centered data regime, such as a rising need for qualitative evaluation methods or the broader range of job responsibilities within this role. These broader responsibilities might include, for example, creating new architecture to interpret data, or developing adjudication methods for human-labelled data.

\subsection{Techniques and Tools}\label{sec:related_tools} There have also been existing inquiries into the techniques and tools that practitioners use. Many data science workers interact with data in tabular formats, using tools such as Google Sheets or Microsoft Excel \cite{birch2018}. They may also writing code to perform custom analyses, commonly by using Python scripts or notebooks such as Google Colab or Jupyter \cite{chattopadhyay, kery, tabard, kery-python}.

As large language models have become more salient, the space of applicable techniques and tools has increased. The field of explainable AI (XAI) \cite{das, danilevsky} has yielded new explainability \cite{tcav, lime} and visualization techniques for natural language processing. These techniques can be packaged into frameworks and tools \cite{kandel,agarwal2020, ashtari2023}, such as Language Interpretability Tool \cite{lit}, What-If Tool \cite{whatiftool}, and AI Fairness 360 \cite{bellamy} among many others \cite{modeltracker, arawjo2023chainforge, ming,  hohman2018visual, la2023state}. However, these LLM-focused tools are relatively recent, and there is a lack of existing research assessing the extent of their adoption across industry and academia.

\newpage
\subsection{Curation Trends}\label{sec:curation_trends}
Datasets relevant to LLM development have become increasingly composed of smaller, curated subsets that target address specific concerns, such as safety and fairness \cite{mehrabi, wagstaff}. The focus is increasingly on data quality \cite{sambasivan} rather than quantity \cite{durrant}, though quantifying the criteria for data quality is an open problem \cite{herman, doshi}.

\section{Retrospective analysis}
To define the role of \textit{data practitioner}, we conducted a retrospective analysis of teams working on developing LLMs at Google. This company's organizational structure is uniquely positioned to support a broad survey of the landscape as the technology stack is vertically-integrated \cite{harrigan1985vertical}; that is, the relevant tooling, infrastructure, modeling, evaluation, and research are primary developed in-house. For example, Google has infrastructure teams that build custom software to deploy ML experiments on computational resources, tooling teams that create applications for interpreting model outputs, data teams that source and clean human data, modeling teams that improve LLM models across different modalities, and safety teams that focus on enforcing policies and model quality. 

Using company-internal organizational charts and employee directories, we identified projects associated with the development of the company's core LLMs. We also conducted a meta-review of company-internal user studies around evaluating tools for data exploration. Applying a grounded theory methodology \cite{corbin}, we inductively applied a relational content analysis and synthesized common themes to develop a framework around data practitioning.

\subsection{Defining the Dataset Practitioner}\label{retro:who}
The \textit{dataset practitioner} interacts with unstructured, text-based data for the purpose of developing large language models. The practitioner's day-to-day work can cover a broad range of tasks traditionally defined in roles such as software developer, machine learning engineer, data scientist, research scientist, product manager, or product counsel. The practitioner may prioritize these responsibilities concurrently, or switch gears along the model development lifecycle. They may do any of the following representative tasks:

\begin{itemize}
    \item Curating a new dataset from scratch
    \item Creating a new benchmark dataset
    \item Cleaning a dataset by removing or fixing bad examples
    \item Analyzing a dataset (feedback, comments, etc) to find trends
    \item Understanding what bias issues might exist in the dataset
    \item Making a go/no go decision on whether to use a dataset to train a model
    \item Debugging a specific model error by finding relevant data
    \item Finding ways to improve models, try different datasets, and compare model results
    \item Identifying key metrics to define ``quality'' for a use case
\end{itemize}

Next, we give examples of datasets that they may explore. The term ``dataset'' traditionally implies static and well-curated data; we expand this notion to include \textit{any} set of text examples, which may come from a variety of provenances (e.g. scraped, synthetically generated, curated by experts). We categorize these broadly:

\newpage 
\begin{enumerate}
\item \textbf{Training datasets}
\begin{itemize}
    \item \textbf{Pre-training data}: LLMs are pre-trained on huge amounts of data from webscrapes, books, and other giant corpora. The curation of these datasets is hugely impactful on the model's performance \cite{longpre2023pretrainers}.
    \item \textbf{SFT and RLHF data}: Supervised Fine-Tuning (SFT) and Reinforcement Learning from Human Feedback (RLHF) datasets are used to refine pre-trained LLMs~\cite{team2023gemini, openai2023gpt}. They are significantly smaller and more specialized than pre-training data, and can be used from an open-ended generation model to a specific use case---most notably, the chatbot interface that many productionized LLMs employ. LLMs can be fine-tuned for other specific products and use cases as well.
\end{itemize}

\item \textbf{Datasets involved in model evaluation}
\begin{itemize}
    \item \textbf{Benchmark evaluation data}: Benchmark datasets are created to test specific functionalities or behaviors of the model. One notable category of these are \textbf{safety benchmarks}, which test the model's ability to adhere to company policies and safety standards on concepts such as toxicity, hallucination, etc.
    \item \textbf{Model outputs}: Model outputs can be evaluated outside of the context of a specific benchmark. Side-by-side analysis of model outputs may be conducted against golden sets or outputs from a baseline model \cite{kahng2024llm}.
    \item \textbf{Outputs of in-context learning:} These are a specific subset of model outputs. In-context learning has allowed users to create new models with no golden data at all. These may then be evaluated by analyzing the outputs from multiple runs of a prompt.
    \item \textbf{Conversational data}: User interactions with LLM-based chatbots can be used to evaluate LLMs in the wild.
\end{itemize}
\end{enumerate}

\section{Qualitative Study} 

\subsection{Participants}
Using our updated definition, we recruited ten dataset practitioners (N=10) within Google for our study.\footnote{Note that Reif et al. \cite{reif2024automatic} uses the same participant sample.} We selected these participants with the criteria that their current work involves interacting with datasets for the purposes of developing large language models, and prioritized sampling participants from a variety of concentrations and backgrounds. These participants and their primary focus areas (tooling, modeling, or evaluation) are listed in \textit{Table \ref{table:participants}}. We validated our observation from \textit{Section \ref{retro:who}} that the domains of their work are fluid; participants who identified in one domain during our recruiting cycle demonstrated experience in many adjacent areas within the interview. For example, a practitioner formerly focused on modeling shifted priorities to safety and fairness evaluation as their models became more scrutinized and regulated, and two tool-builders reported being driven to build tooling to address their own unmet needs in modeling.

\begin{table}[!t]
\begin{tabular}{|l|l|l|}
\hline
\rowcolor[HTML]{C0C0C0} 
\textbf{Domain} & 
\textbf{Participant ID} & 
\textbf{Focus Area} \\ \hline
                             & T1 & Tools for data annotation    \\ \cline{2-3} 
                             & T2 & Tools for data curation      \\ \cline{2-3} 
                             & T3 & Tools for data understanding \\ \cline{2-3} 
\multirow{-4}{*}{Tooling}    & T4 & Pipeline infrastructure      \\ \hline
                             & M1 & Data curation.               \\ \cline{2-3} 
                             & M2 & Model architecture           \\ \cline{2-3} 
\multirow{-3}{*}{Modeling}   & M3 & Model refinement             \\ \hline
                             & R1 & Robustness and abuse         \\ \cline{2-3} 
                             & R2 & Unsafe and sensitive content \\ \cline{2-3} 
\multirow{-3}{*}{Evaluation} & R3 & Annotator ethnography        \\ \hline
\end{tabular}%
\caption{Study participants and their current focus areas, grouped by domain.}
\label{table:participants}
\end{table}
% \vspace{-30pt}

\newpage
\subsection{Interview Protocol}

Following recruitment and an informed consent process, we conducted semi-structured, one-on-one interviews with participants over video conferencing. Each 30-minute interview covered the following topics:
\begin{enumerate}
    \item \textit{Understanding the use case:} Background, use case, product impact, research questions
    \item \textit{Tools and techniques}: Awareness and usage of existing tools and pipelines, decision making, advantages and limitations, statistical and visual interpretability methods
    \item \textit{User challenges}: Bottlenecks, unaddressed concerns 
\end{enumerate}

We curated the interview topics from prior contextual inquiries and protocols from similar research studies in defining data work \cite{kaur, li, wang2019}. By following a similar interview protocol, we hope to isolate the specific challenges faced in LLM-development.

We synthesized our findings through a thematic analysis \cite{braun}. Each interview was de-identified, transcribed, broken into excerpts, and coded. Thematic elements, behaviors, and representative quotes in this paper are saturated \cite{Guest2006, Ando2014}, with a code repeated in at least three distinct transcriptions. 

\begin{figure*}[!t]
\begin{tabular}{cc}
  \includegraphics[width=0.9\textwidth]{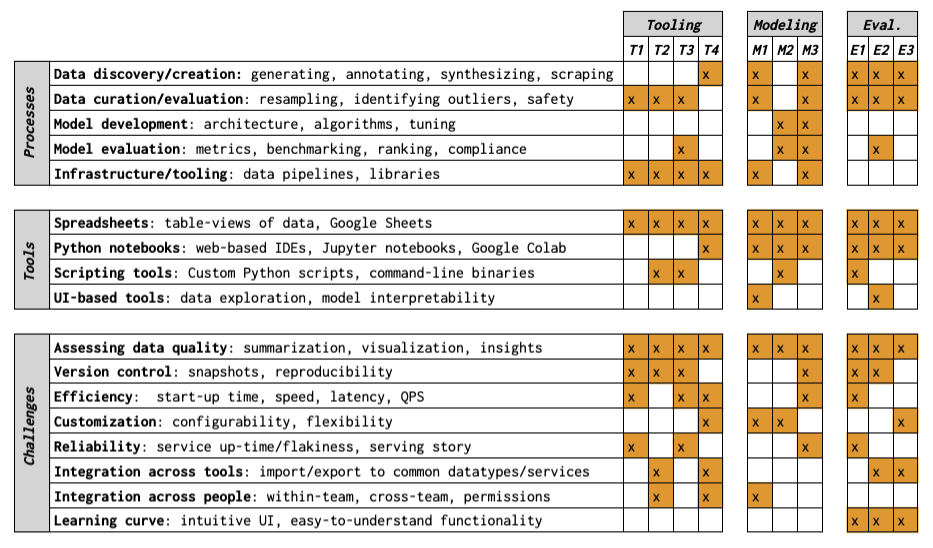}
\end{tabular}
\captionof{table}{This matrix categorizes our findings (inspired by Kandel et al. \cite{kandel}). An `x' in the cell indicates that a participant mentioned this specific topic in their interview. Topics are grouped by \textit{Processes, Tools,} and \textit{Challenges}, and participant are grouped by their domain from \textit{Table \ref{table:participants}}. All participants mentioned interacting with spreadsheets and cited data quality as a challenge in their work.}
\Description[A matrix categorizing findings.]{There are 10 columns, one for each participant. Each row is a specific theme that was discovered and coded. All participants mentioned interacting with spreadsheets and cited data quality as a challenge in their work.}
\label{table:matrix}
\end{figure*}
% \vspace{-12pt}

\subsection{Findings}

\subsubsection{\textbf{Participants prioritize data quality.}}\label{result:quality}

Corroborating the prior work described in Section \ref{sec:curation_trends}, we find that data quality---defining, finding, and identifying high-quality data---was unanimously the biggest user challenge and priority across all use cases (\textit{Table \ref{table:matrix}, Challenges}).

    \begin{quote}
        \textit{ Data, historically, has been around volume rather than quality.. we've had this big paradigm shift. }\hspace{1em plus 1fill}---T2
    \end{quote}
    \begin{quote}
        \textit{``Quality is the big obstacle… [You need] a lot of high-quality data... there's no shortcut.''}\hspace{1em plus 1fill}---E1
    \end{quote}

Although data quality has always remained an important priority for data scientists, these concerns were addressable through tasks such as data cleaning \cite{muller} or feature engineering \cite{crisan}. In the context of generative modeling, the evaluation metrics and consensus frameworks are less straightforward.

\newpage
\subsubsection{\textbf{However, practitioners rely largely on their own intuition to validate this data quality.}}

All participants reported that they would evaluate their data by scanning it visually in spreadsheet form; that is, they would look at a handful of examples.

   \begin{quote}
        \textit{``I’ll read the first 10 examples, and then maybe some in the middle.''}\hspace{1em plus 1fill}---E1
    \end{quote}

    \begin{quote}
        \textit{``I eyeball data.. It’s all my own intuition and kind of individually spot checking examples.''}\hspace{1em plus 1fill}---M2
    \end{quote}
    
Participants cited efficiency, customization, a short learning curve, and ease-of-sharing as reasons for their reliance on spreadsheets (\textit{Table \ref{table:matrix}, Challenges}).\footnote{Interestingly and consistent with similar user studies, our participants emphasized that their reliance on visual inspection of spreadsheets were their \textit{own} behaviors and not best practices. They suggested that other practitioners likely used more sophisticated tooling \cite{pennebaker}.} While these factors align with prior research on spreadsheet usage \cite{birch2018}, the \textit{ease-of-sharing} factor may particularly encourage practitioners to use spreadsheets for LLM development. Unlike the data analysts in Kandel et al. \cite{kandel}, who collaborated with ``hacker''-types with scripting and coding proficiency, our participants reported needing to share data with a larger and more diverse set of stakeholders, such as directors and legal teams, to review high-stakes safety fine-tuning datasets.

\subsubsection{\textbf{Or, practitioners will run custom analyses.}}

Seven of the nine participants mentioned also writing custom code in Python notebooks to explore their data, and in one instance even to train production models. Participants liked the customization of these notebooks \cite{kery}, yet cited reliability, setup, efficiency, code management as pain points (\textit{Table \ref{table:matrix}, Challenges}), validating results from other studies on Python notebook usage \cite{kery-python, chattopadhyay, kery, tabard}. 

The efficiency concerns around long-running computations in Python notebooks \cite{chattopadhyay} may be further exacerbated as LLMs require more computational power; participants mentioned that ``getting model servers up and running takes forever'' (R1), ``my queries [to LLM APIs] take a while'' (E1), and they wished they had ``infinite QPS (Queries Per Second) [for their LLM API]'' (R2).

\subsubsection{\textbf{Practitioners recognize the confirmation biases in their exploration practices.}}

The majority---if not all---of the data exploration is being done between visual inspection in spreadsheets and custom logic in Python notebooks, allowing the practitioner to look at whatever they would like. This degree of freedom exacerbates cognitive bias \cite{herman, pirolli, gilpin, caliskan}; for example, Miller et al. \cite{miller} mentions that ``explainable AI uses only the researchers' intuition of what constitutes a `good' explanation.'' Indeed, our participants admit to this confirmation bias in their practices:

    \begin{quote}
        \textit{``“I eyeball that things make sense [in the data].''}\hspace{1em plus 1fill}---M2
    \end{quote} 

In fact, model developers reported that they did not look at training data unless their model outputs were surprising.

    \begin{quote}
        \textit{``When the data is passed to the modeling side, we assume that the data team has fixed everything. Unless we train and it doesn't look right, then we’ll [look at the data] and give the data team that feedback.''}\hspace{1em plus 1fill}---M3
    \end{quote}

\newpage
\subsubsection{\textbf{Participants have not converged upon other tools.}}
Apart from Google Sheets and Python notebooks like Colab, no other tools garnered consensus among practitioners. Some practitioners employed additional methods, such as running a binary for calculating safety and toxicity thresholds, kicking off a pipeline to automatically classify their data, and using an user interface to visualize embeddings. However, these practices were not prevalent in our sample.

    \begin{quote}
        \textit{``Everyone is using a different thing, and getting everyone on the same page is really difficult.''}\hspace{1em plus 1fill}---M1
    \end{quote}

The lack of alignment in tooling presents an organization challenge. As training datasets are increasingly composed of smaller datasets to leverage the expertise of specific subteams, greater collaboration across groups is necessary. This can lead to increased friction in adopting new tools and exploration patterns \cite{kandel}, as stakeholders and collaborators must transition to new tooling simultaneously, or migrate in a manner that preserves data sharing capabilities.

    \begin{quote}
        \textit{``With the new generative data--- Many people are contributing with many different lenses. In practice, these [subsets] get built by random teams, they get added and nobody really reviews it because you can't.''}\hspace{1em plus 1fill}---T4
    \end{quote}

\section{Discussion}

The reason why practitioners have not aligned on alternative tooling is not obvious. Practitioners across all domains recognize that there is a gap in the workflow:

    \begin{quote}
        \textit{``Not having an easy-to-use-tool is a major bottleneck… Every time [that I make changes to data], I have to write a custom colab to ingest the new fields.''}\hspace{1em plus 1fill}---M2
    \end{quote}

    \begin{quote}
        \textit{``There are no helpful tools from a qualitative researcher’s perspective. I jump between spreadsheets, a CSV file and a colab… The long story short is that we haven't really found a very useful tool for this.''}\hspace{1em plus 1fill}---E3
    \end{quote}

    \begin{quote}
        \textit{``Right now, if you want to curate high-quality data, you go through [each point] manually as an expert, which is not scalable [for] thousands of examples.''}\hspace{1em plus 1fill}---T2
    \end{quote}
    
Practitioners are aware of and have tried the existing tools in this space. They are aligned on the properties that they want out of this tool (\textit{Table \ref{table:matrix}, Challenges}), and these requests are being communicated to tooling teams:
    \begin{quote}
        \textit{``The kinds of requests we tend to get nowadays are about larger-scale dataset management, like mixture building. When you have a big selection, reviewing 10,000 rows is not what you want to do \ldots That is much more amenable to summary review.''}\hspace{1em plus 1fill}---T1
    \end{quote}
 
In response, tooling teams are evaluating and building tools to address these requests \cite{lit, modeltracker, whatiftool}. So, why is there a lack of alignment? We discuss hypotheses posed by two different domains of practitioners.

\subsubsection{\textbf{The toolmakers' hypothesis: the world is new.}}

When tool developers (T1-T4) described exploration workflows, they explained that there was a lack of alignment because the field is new:

    \begin{quote}
        \textit{``The pace is very frenetic right now.. tools are fast-changing\ldots''}\hspace{1em plus 1fill}---T1
    \end{quote}
    
    \begin{quote}
        \textit{``There's been a big step function in the NLP world.. it just takes a while to figure out what tools people need and what all use cases.''}\hspace{1em plus 1fill}---T2
    \end{quote}

Two observations from our interviews may support this claim. First, practitioners are using spreadsheets. Perhaps in the absence of a ground truth for unstructured data, practitioners prefer to rely on their own intuition. Similarly, without a definitive framework for qualitative data exploration, practitioners are sticking to the tools they know. Adopting new practices takes effort (see \textit{Table \ref{table:matrix}, Challenges > Learning curve}), and spreadsheets have been tried-and-true from the previous state-of-the-art when visually spot-checking data and conducting statistical analyses were sufficient. 

Second, our participants described a landscape where there was a lack of alignment \cite{doshi, gilpin} across multiple topics such as objectives, metrics, and benchmarks, suggesting that the field and its principles are still emerging. The following are representative quotes from participants:

\begin{itemize}
    \item \textbf{Data quality}: 
        \begin{itemize}
            \item T1, on LLM prompts: \textit{``There's so many competing definitions of prompt quality\ldots  it's a research north star that happens to be a major product priority. How can we improve this extremely important data set?'}
            
            \item M1, on training data:
            \textit{``The quality of data is subjective; a lot of people disagree\ldots one person thinks it's really high-quality data, but there's no objective.''}
            
            \item T3, on evaluation data: \textit{``There's not a framework for evaluating [data].. in a perfect world, there is well-articulated behavior (tone, subject matter, objective results)..''}
        \end{itemize}
        
    \item \textbf{Metrics}: 
        \begin{itemize}
            \item M1: \textit{``[Consider] search rankings\ldots what makes for a good benchmark, how do we come to an agreement?''}
            
            \item E1: \textit{``If you're doing simple classification, it's easy to measure accuracy or precision or recall. But with generative models, evaluation is very subjective. Even the output of the model is subjective, so then, what's going into the model- it's really hard to say, is this better or worse?''}
        \end{itemize}

    \item \textbf{Safety}:
        \begin{itemize}
            \item T2: \textit{``Think about safety data curation\ldots people can't agree on criteria, let alone apply that criteria at scale.''}\hspace{1em plus 1fill}
        \end{itemize}
        
    \item \textbf{Communication}: 
        \begin{itemize}
            \item T3: \textit{``What [data practitioners are] actually doing and what they communicate that they need are two very different things. What are they actually trying to do?''}
        \end{itemize}
\end{itemize}

This lack of alignment is amplified as teams collaborate more closely \cite{nahar2022,zhang2020}. Even if one team in the development pipeline identifies their quality evaluation parameters, there needs to be further agreement at the inter-team level. 

\subsubsection{\textbf{The model developers' hypothesis: there's no tool that works for my use case.}}

Modeling and evaluation practitioners speculated that alignment was unlikely due to custom needs and requirements (\textit{Section \ref{retro:who}}).

\begin{quote}
    \textit{``I think why [a spreadsheet is] so universal is that it's so basic.. you can customize it to give this affordance that other tools may not give you.. it's simple.''}\hspace{1em plus 1fill}---E1
\end{quote}

\begin{quote}
    \textit{``We have tried so many [tools]. These tools are limiting is because they offer you exploration on only one aspect of [the data]\dots For me, they're too specific.''}\hspace{1em plus 1fill}---M2
\end{quote}

Interestingly, when asked about the custom requirements for their use cases, practitioners listed similar requirements, which suggest that there may be opportunities for shared methods and evaluation frameworks. Some of these requirements include:
\begin{itemize}
    \item Summarizing salient features of a dataset and identifying the corresponding data slices (6 participants)
    \item Ensuring safety of outputs/respecting toxicity thresholds (4 participants)
    \item Evaluating numeric distributions on text/token length (3 participants)
\end{itemize}

It is likely the case that both the toolmakers' and model developers' hypotheses are true to some extent. There may be select opportunities for alignment as the field matures, and there are likely other problems that will require custom solutions. For example, there are specific tools being developed to address challenges that persist across datasets, such as safety and toxicity classification \cite{bellamy}.

\section{Conclusions and Future Work}

In this study, we aimed to identify the needs of those who are exploring unstructured, text-based datasets for the purpose of developing LLMs. To define this population of \textbf{dataset practitioners}, we conducted a retrospective analysis on teams working on LLM development. We then interviewed a broad cross-section of these practitioners to better understand their use cases and challenges.

Through our retrospective analysis, we found that the dataset practitioner takes on a fluid role that is not well-defined in current literature on data workers. We hope that our contribution of defining this population and their use cases will enable the HCI community to better assess and support their needs.

In our interviews, we found that data quality is unanimously the top priority, but quality is subjective. Further research should explore what data quality means in different contexts, and how the same data can be high-quality or low-quality depending on the situation and perspective. Clarifying subjectivity across conceptual frameworks, evaluations, and workflows in this domain remains a top priority, potentially achieved through standardizing metrics (e.g. toxicity, distributions of relevant safety features, data diversity) and evaluation criteria.

Two primary data exploration patterns emerge: visually inspecting data in spreadsheets, which lacks scalability, and crafting custom analyses in Python notebooks, which is high-effort. Both practices are susceptible to confirmation bias. However, the community has yet to reach a consensus on alternative best-practices to for data exploration, possibly due to the nascent nature of the field or the custom needs of the practitioners. There are opportunities to determine the specific areas where prioritizing either flexibility or specificity is most beneficial; these opportunities can be addressed by formalizing evaluation frameworks in the evolving landscape, and developing flexible tooling for custom analysis. 

\vspace{10pt}
\begin{quote}
    \textit{``There's a fundamental chicken and egg problem\ldots there's no tooling so people don't use tooling so tooling doesn't develop.''}\hspace{1em plus 1fill}---T2
\end{quote}
\newpage
%%
%% The acknowledgments section is defined using the "acks" environment
%% (and NOT an unnumbered section). This ensures the proper
%% identification of the section in the article metadata, and the
%% consistent spelling of the heading.
\begin{acks}
The authors wish to thank our study participants and Google's People + AI Research Team (PAIR), especially James Wexler and Michael Terry.
\end{acks}

%%
%% The next two lines define the bibliography style to be used, and
%% the bibliography file.

\bibliographystyle{ACM-Reference-Format}
\bibliography{references.bib}

%%
%% If your work has an appendix, this is the place to put it.
\
\appendix

\end{document}